%% file: main.tex
\pdfoutput=1
\documentclass[11pt]{article}
\DeclareUnicodeCharacter{02BC}{'}

\usepackage[final]{acl}

\usepackage{times}
\usepackage{latexsym}
\usepackage[T1]{fontenc}
\usepackage[utf8]{inputenc}
\usepackage{microtype}
\usepackage{inconsolata}
\usepackage{tikz-dependency}
\usepackage{graphicx}
\usepackage{booktabs}
\usepackage{amsmath}
\usepackage{amssymb}
\usepackage{makecell}
\usepackage{bbm}
\usepackage{paralist}
\usepackage{multicol}
\usepackage{multirow}
\usepackage{subcaption}
\usepackage{xurl}
\usepackage{hyperref}

\title{Efficient Environmental Claim Detection with Hyperbolic Graph Neural Networks}

\author{
\textbf{Darpan Aswal}$^{1,2}$,
\textbf{Manjira Sinha}$^{3}$ \\
$^1$Department of Computer Science, Université Paris-Saclay \\
$^2$MICS, CentraleSupélec, Université Paris-Saclay \\
$^3$TCS Research, India \\
\footnotesize{\textbf{Correspondence}: \href{mailto:darpan.aswal@universite-paris-saclay.fr}{darpanaswal@gmail.com}}
}

\begin{document}
\maketitle
\begin{abstract}
Transformer based models, especially large language models (LLMs) dominate the field of NLP with their mass adoption in tasks such as text generation, summarization and fake news detection. These models offer ease of deployment and reliability for most applications, however, they require significant amounts of computational power for training as well as inference. This poses challenges in their adoption in resource-constrained applications, especially in the open-source community where compute availability is usually scarce. This work proposes a graph-based approach for Environmental Claim Detection, exploring Graph Neural Networks (GNNs) and Hyperbolic Graph Neural Networks (HGNNs) as lightweight yet effective alternatives to transformer-based models. Re-framing the task as a graph classification problem, we transform claim sentences into dependency parsing graphs, utilizing a combination of word2vec \& learnable part-of-speech (POS) tag embeddings for the node features and encoding syntactic dependencies in the edge relations. Our results show that our graph-based models, particularly HGNNs in the poincaré space (P-HGNNs), achieve performance superior to the state-of-the-art on environmental claim detection while using up to \textbf{30x fewer parameters}. We also demonstrate that HGNNs benefit vastly from explicitly modeling data in hierarchical (tree-like) structures, enabling them to significantly improve over their euclidean counterparts. We make our implementation publicly available~\footnote{\href{https://github.com/darpanaswal/ecd-hgnn}{https://github.com/darpanaswal/ecd-hgnn}}.
\end{abstract}

\input{latex/introduction}
\input{latex/related}
\input{latex/methodology}
\input{latex/experimental_setup}
\input{latex/results}
\input{latex/discussion}
\input{latex/conclusion}
\input{latex/limitations}

\bibliography{main}
\clearpage
\appendix
\input{latex/supplementary}

\end{document}

%% file: latex/introduction.tex
\section{Introduction}
Claim verification and claim detection~\cite{soleimani2020bert, levy2014context} are complex NLP tasks that involves the detection of fake claims using facts as well as contextual information within the given claims. Often, these claims exhibit hierarchical and nested information such as conditional statements~\cite{kargupta2025beyond}. Environmental claim detection~\cite{stammbach2022environmental} involves additional elements from greenwashing~\cite{de2020concepts} that are often used by corporations to promote products and mislead customers.

Recent work for claim detection, similar to many industrial NLP applications~\cite{chkirbene2024large}, has predominantly relied on transformer-based architectures~\cite{ni2024afacta}. However, this reliance on these massive, black-box models presents two issues. First, they require large-scale computational resources which makes them economically and environmentally expensive, leaving behind a large carbon footprint~\cite{faiz2023llmcarbon}. Second, their lack of interpretability~\cite{lin2023generating} is a significant issue in high stakes domains like claim verification, where explaining a classification is equally important as the classification itself~\cite{atanasova2024generating, brundage2020toward}. The increasing scrutiny on sustainability claims further necessitates interpretability and computational efficiency in models.

To address these challenges of cost and interpretability, we propose a lightweight framework for graph-based claim detection.  We re-frame the problem of environmental claim detection as a graph classification task, explicitly modeling the syntactic and hierarchical structure of sentences using dependency parsing graphs~\cite{nivre2010dependency} with word embeddings for node features. This representation provides a natural fit for Graph Neural Networks (GNNs)~\cite{wu2020comprehensive} which are designed to learn from such structured data. Compared to transformers, our approach offers an interpretable approach to syntactic and semantic learning while significantly reducing computational overhead~\cite{feng2025grapheval, li2025hybrid, peng2021hyperbolic}. Furthermore, given the tree-like nature of dependency graphs, we investigate Hyperbolic Graph Neural Networks (HGNNs)~\cite{zhou2023hyperbolic}, a geometric learning architecture particularly suited to such hierarchically structured data. The research questions for the study are as follows.

\noindent\textbf{RQ1. }Can graph-based models match SOTA performance for environmental claim detection while using just a fraction of the compute as that of LLMs?

\noindent\textbf{RQ2. }Can syntactically enriched explicit hierarchical modeling of NLP tasks advantage hyperbolic models over their euclidean counterparts?

%% file: latex/related.tex
\begin{figure*}[!h]
    \centering
    \includegraphics[width=1\textwidth, keepaspectratio]{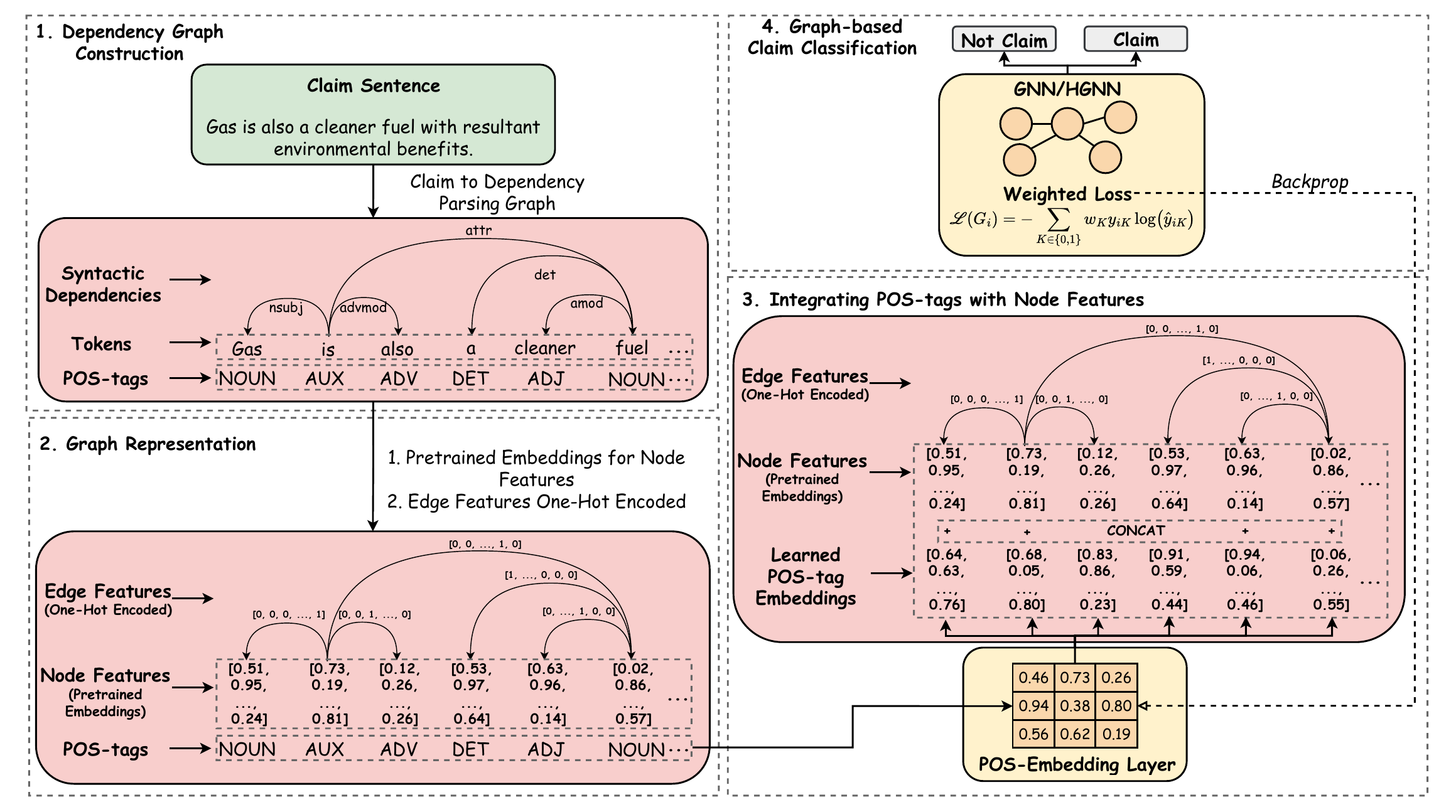}
    \caption{Overview of the Graph-based Claim Detection Pipeline. Step 1: Claim sentence to dependency graph conversion. Steps 2: Dependencies are one-hot encoded as edge features. Node features are initialized with pretrained embeddings. Step3: Node features are concatenated with POS-tag embeddings learned by embedding layer. Step 4: Graph classification using a GNN/HGNN architecture trained with a weighted loss function.}
    \label{fig:methodology}
\end{figure*}

\section{Related Work}
The proliferation of misinformation on social media has shown the need for automated fact-checking and verification systems~\cite{aimeur2023fake}. Fake news detection aims to classify entire articles or posts as credible or fake~\cite{shu2017fake}, often involving analyzing of multiple signals such as textual content and writing style~\cite{przybyla2020capturing}. While early approaches relied on feature engineering and machine learning methods~\cite{khanam2021fake}, recent work relies on transformer models for fake news detection~\cite{yi2025challenges}.

Claim detection and verification offer a more detailed approach to fact-checking. Claim detection~\cite{levy2014context} focuses on identifying factual statements within larger texts and separating them from non-factual ones. Claim verification~\cite{soleimani2020bert} on the other hand assesses the accuracy of detected claims using evidence and facts from trusted sources. While fact checking is widely utilized for social-media content~\cite{wasike2023you}, these methods have been applied to specific, high-stakes topics such as verification of climate-related claims~\cite{diggelmann2020climate} and analyzing contrarian~\cite{coan2021computer} or fake claims about climate change~\cite{al2021twitter}. Environmental claim detection~\cite{stammbach2022environmental} is one such specialized sub-domain of fact verification research. Specifically, it deals with greenwashing~\cite{de2020concepts} -- the corporate form of misinformation -- which involves using vague or misleading language to create an exaggeratedly positive public image of a company's environmental credentials. 

Large Language Models (LLMs)~\cite{naveed2025comprehensive} are transformer-based models~\cite{lin2022survey} pre-trained on vast amounts of text data which enables them to achieve state-of-the-art performance in downstream tasks such as sentiment analysis, machine translation and named entity recognition~\cite{miah2024multimodal, zhang2023prompting, yan2019tener}. The application of these models has evolved from fine-tuning~\cite{wu2025llm} task specific models such as BERT and RoBERTa~\cite{soleimani2020bert, stammbach2022environmental}, to in-context learning~\cite{dong2022survey} with modern, multi-billion parameter models. While powerful, the high computational costs~\cite{faiz2023llmcarbon} and lack of interpretability~\cite{lin2023generating} of these models pose challenges for wide-scale adoption.

Graph Neural Networks (GNNs)~\cite{wu2020comprehensive} offer an alternative learning paradigm by operating on structured-data. Prior work has utilized GNNs to explicitly model hierarchical and relational dependencies~\cite{mi2020hierarchical}, making graph structures such as constituency parsing~\cite{li2020empirical} and dependency parsing graphs~\cite{nivre2010dependency} a natural fit for representing sentence structures in NLP tasks. These models can integrate rich semantic information from word embeddings, knowledge graphs, and even sentence embeddings from pre-trained language models~\cite{mikolov2013efficient, opdahl2022semantic, li2020sentence}. Geometric deep learning~\cite{bronstein2017geometric} generalizes these models to non-euclidean spaces~\cite{coxeter1998non}. Extending GNNs, Hyperbolic GNNs~\cite{zhou2023hyperbolic}, are particularly well suited to model hierarchically structured data such as dependency parsing graphs.

%% file: latex/methodology.tex
\section{Methodology}
We begin our experimentation by transforming a dataset $D = \{c_1, c_2, \dots, c_N\}$ of $N$ environmental claims into a corresponding set of dependency parsing graphs $G = \{G_1, G_2, \dots, G_N\}$, converting each claim $c_i$ into a unique graph structure 
\[
    G_i = (V_i, E_i)
\]
where $V_i$ is the graph's set of vertices (or nodes) and $E_i$ is its set of edges.
\subsection{Dependency Graph Construction}
For each claim $c_i$ in the dataset, we generate a directed dependency graph using spaCy's built-in DependencyParser. Claim $c_i$, which is a sequence of tokens $t_i = \{t_i^{(1)}, t_i^{(2)}, \dots t_i^{(k)}\}$ is mapped to its corresponding graph $G_i = (V_i, E_i)$ where the vertex set $V_i = \{v_1, \dots v_k\}$ represents the tokens, and the edge set $E_i$ represents the syntactic dependencies between them. A directed edge $(v_h, v_j) \in E_i$ exists if the token $t_h$ is the syntactic head of token $t_j$. Each edge is labeled with its dependency type $d \in D$, where $D$ is the set of all 45 unique dependency relations present in the dataset. We utilize the following node and edge attributes from the dependency graphs~\footref{common_foot}.
\begin{compactitem}
    \item \textbf{Token text}: Represented as the graph's nodes; corresponds to tokens in the claim sentences.
    \item \textbf{Dependency relation}: Specifies the type of syntactic dependency between a token and its head. Describes how the token relates to its syntactic governor.
    \item \textbf{Token head}: Also represented as the graph nodes, it identifies the governor token for a given dependent token.
    \item \textbf{Token Part-Of-Speech (POS) tag}.
\end{compactitem}

\subsection{Graph Representation}
To prepare the graphs for the GNN models, we define the node and edge feature representations.
\subsubsection{Node Features}
Each node $v \in V_i$ is associated with a feature vector $x_v \in \mathbb{R}^{d_{node}}$. For this vector, we utilize word2vec \citep{mikolov2013efficient}, a pre-trained word embedding model. $x_v = W_e(\text{token}(v))$, where $W_e$ is the word2vec embedding lookup matrix and the token$(v)$ is the word corresponding to node $v$.

% Other embedding approaches include GloVe~\cite{pennington2014glove}, fastText~\cite{bojanowski2017enriching}, and transformer embeddings~\cite{li2020sentence}, each with their own set of advantages for extracting semantic information from text input. However, a comparison among different embedding models lies outside the purpose of our study.

\subsubsection{Edge Features}
The syntactic dependency type of each edge, corresponding to one of the 45 unique relations in the dataset, is encoded into a feature vector. For an edge $e=(v_h, v_j)$, its feature vector $e_{hj} \in \mathbb{R}^{|D|}$ is a one-hot encoding of its dependency type $d(e)$. 

\subsection{Integrating POS-tags with Node Features}
Next, we augment the node features with the POS-tags. Let $\mathcal{P}$ be the set of all unique POS-tags in the dataset. We 
introduce a learnable embedding matrix $W_p \in \mathbb{R}^{|\mathcal{P}| \times d_{pos}}$, 
where $d_{pos}$ is the dimension of the POS-tag embeddings. This layer is trained with the GNN model. The final feature vector 
for a node $v$, denoted $x'_v$, is the concatenation of its word embedding and its 
learned POS tag embedding:

\[
x'_v = \big[ W_e(\text{token}(v)) \,\|\, W_p(\text{pos}(v)) \big]
\]

\noindent The dimension of this augmented feature vector becomes 
$d'_{node} = d_{node} + d_{pos}$.

\subsection{Weighted Loss for Imbalanced Data}
Lastly, to address the inherent imbalance present in the dataset, we employ a weighted cross-entropy loss function. This strategy assigns a higher penalty to misclassifications of the minority class, thereby encouraging the model to pay more attention to it. The loss for a single graph $G_i$ with true one-hot label $y_i$ and predicted probabilities $\hat{y}_i$ is defined as:
\[
    \mathcal{L}(G_i) = -\sum_{k=0}^1 w_k . y_{ik}log(\hat{y}_{ik})
\]
The weight for each class $k, w_k$, is calculated as the inverse of its frequency in the training set, effectively balancing the contribution of each class to the overall loss.

\subsection{Graph-based Claim Classification}
The final stage of our pipeline involves classifying the entire graph representation of a claim sentence. The augmented node feature vectors and the edge feature vectors are fed into either a GNN or an HGNN model which provides the final classification for the claim sentences, classifying them into two possible categories -- `Claim' and `Not Claim'.

%% file: latex/experimental_setup.tex
\begin{table*}[!h]
\centering
\normalsize
\begin{tabular}{lcccc|cccc}
\toprule
\textbf{Model} & \multicolumn{4}{c|}{\textbf{dev}} & \multicolumn{4}{c}{\textbf{test}} \\
 & pr & rc & F1 & acc & pr & rc & F1 & acc \\
\midrule
DistilBERT  
& \textbf{77.5} & \textbf{93.9} & \textbf{84.9} & 91.7  & 74.4 & \textbf{95.5} & 83.7 & 90.6 \\
ClimateBERT 
& 76.9 & 90.9 & 83.3 & 90.9 & 76.5 & 92.5 & 83.8 & 90.9 \\
RoBERTa\textsubscript{base}  
& 74.7 & \textbf{93.9} & 83.6 & 90.6 & 73.3 & 94.0 & 82.4 & 89.8 \\
RoBERTa\textsubscript{large}
& \textbf{80.5} & \textbf{93.9} & \textbf{86.7} & \textbf{92.8} & \textbf{78.5} & 92.5 & \textbf{84.9} & \textbf{91.7} \\
\bottomrule
\end{tabular}
\caption{Results reported by~\cite{stammbach2022environmental} on their ECD-dataset.}
\label{tab:baselines}
\end{table*}
\section{Experimental Setup}
\subsection{Dataset}
We utilize the Environmental Claim Detection (ECD) dataset~\cite{stammbach2022environmental}, a dataset comprised of environmental claims extracted from various corporate communications of publicly listed companies, including sustainability reports, earnings calls, and annual reports.
While the authors initially collected 3,000 sentences, they removed samples with tied annotations, reporting results on the filtered dataset of 2,647 samples. We use this same 2,647-sample dataset for all our experiments to ensure a direct comparison. The dataset is imbalanced, with 665 sentences (25.1\%) labeled as claim statements and 1,982 sentences (74.9\%) labeled as not claim statements.

\subsection{Models}
We conduct our analysis with Euclidean and Hyperbolic GNN architectures. For training our models, we utilized the HGNN toolkit from \citep{liu2019hyperbolic}. We experiment with the two standard models of hyperbolic space -- the Poincaré Ball~\cite{nickel2017poincare}, which represents the hyperbolic space inside a unit disk and the Lorentz Hyperboloid Model~\cite{nickel2018learning} which embeds the space on a hyperboloid~\cite{reynolds1993hyperbolic} in a higher-dimensional Minkowski space~\cite{naber2012geometry}. Our models are trained with a total of 4 GNN layers. The first layer's dimensionality $d_{in} = d_{word2vec} + d_{pos} \text{, where } d_{word2vec} = 300$. The other 3 GNN layers are of dimensionality 256. For training, we utilize the AMSGrad and the Riemannian AMSGrad optimizers for the GNN and HGNN respectively~\footref{common_foot}.

\subsection{Evaluation Metrics}
For evaluating our models, we use five primary metrics to assess their performance on the claim detection task -- Accuracy, Precision, Recall, F1-score, and AUC-ROC~\footref{common_foot}. Given the high imbalance in the dataset, we use the F1-score and AUC-ROC as our primary metrics

%% file: latex/results.tex
\section{Results \& Observations}
\begin{table*}[!h]
\centering
\scriptsize
\begin{tabular}{lccc|ccccc|ccccc}
\toprule
\textbf{Model} & \multicolumn{3}{c|}{\textbf{grid-search parameters}} & \multicolumn{5}{c|}{\textbf{dev}} & \multicolumn{5}{c}{\textbf{test}} \\
 & \makecell{Dropout \\ Rate} & \makecell{POS \\ Embedding \\ Dimension} & \makecell{Class \\ Weights} & pr & rc & F1 & acc & auc & pr & rc & F1 & acc & auc \\
\midrule
GNN 
& 0.1 & -- & --
& \textbf{79.3} & 69.7 & 74.2 & \underline{87.9} & \textbf{0.93} & 78.7 & 71.6 & 75.0 & 87.9 & 0.93 \\
L-HGNN 
& 0.1 & -- & --
& 70.3 & 78.8 & 74.3 & 86.4 & \underline{0.92} & 73.7 & 83.6 & 78.3 & 88.3 & 0.93 \\
P-HGNN 
& 0 & -- & --
& 71.0 & 74.2 & 72.6 & 86.0 & 0.91 & 74.4 & \textbf{86.6} & 80.0 & 89.1 & \underline{0.94} \\

\midrule

GNN-POS 
& 0.3 & 32 & --
& 75.4 & 74.2 & 74.8 & 87.5 & \underline{0.92} & 77.9 & 79.1 & 78.5 & 89.1 & \underline{0.94} \\
L-HGNN-POS 
& 0.1 & 64 & --
& 70.5 & 65.2 & 67.7 & 84.5 & \underline{0.92} & 78.3 & 80.6 & 79.4 & 89.4 & 0.93 \\
P-HGNN-POS 
& 0.3 & 128 & --
& 75.4 & 74.2 & 74.8 & 87.5 & \textbf{0.93} & \textbf{85.9} & 82.1 & \textbf{84.0} & \textbf{92.1} & \textbf{0.95} \\

\midrule

Balanced-GNN 
& 0.1 & -- & [1,1.5]
& \underline{78.6} & 66.7 & 72.1 & 87.2 & \textbf{0.93} & \underline{81.7} & 73.1 & 77.2 & 89.1 & 0.93 \\
Balanced-L-HGNN 
& 0.25 & -- & [0.8,1.6]
& 77.8 & 74.2 & \underline{76.0} & \textbf{88.3} & \textbf{0.93} & 75.7 & 79.1 & 77.4 & 88.3 & 0.93 \\
Balanced-P-HGNN 
& 0.2 & -- & [1,1.5]
& 73.2 & \underline{78.8} & 75.9 & 87.5 & \underline{0.92} & 73.7 & 83.6 & 78.3 & 88.3 & 0.93 \\

\midrule

Balanced-GNN-POS 
& 0.25 & 32 & [0.6678,1.9897]
& 72.9 & 77.3 & 75.0 & 87.2 & \textbf{0.93} & 76.7 & 83.6 & 80.0 & 89.4 & 0.93 \\
Balanced-L-HGNN-POS 
& 0 & 16 & [0.6678,1.9897]
& 73.5 & 75.8 & 74.6 & 87.2 & \textbf{0.93} & 74.0 & \underline{85.1} & 79.2 & 88.7 & \underline{0.94} \\
Balanced-P-HGNN-POS 
& 0.3 & 32 & [0.8,1.6]
& 73.6 & \textbf{80.3} & \textbf{76.8} & \underline{87.9} & \textbf{0.93} & 80.3 & \underline{85.1} & \underline{82.6} & \underline{90.9} & \underline{0.94} \\
\bottomrule
\end{tabular}
\caption{We report precision, recall, F1 score, accuracy and the auc-roc score on the dev and test sets of the ECD dataset. The best performance per split is indicated in bold, the second best is underlined.}
\label{tab:results}
\end{table*}

\noindent To obtain the best performance for each model configuration, we grid-search over the dropout rate, POS-embedding dimension and class weights. Next, we describe our results in detail in relation to the research questions described earlier.

\subsection{HGNNs Match SOTA Performance with upto 30x Fewer Parameters (RQ1.)}
In Table~\ref{tab:baselines}, we first establish the baselines using the results from~\cite{stammbach2022environmental} with 4 transformer models -- DistilBERT, ClimateBERT, RoBERTa\textsubscript{base}, and RoBERTa\textsubscript{large}. While we use F1-score and AUC-ROC as our primary metrics, we include the standard accuracy in our tables solely for a direct comparison with the baseline metrics. In Table~\ref{tab:results}, we see that our graph-based models achieve performance better than or comparable to these state-of-the-art transformers. Firstly, our simplest models -- labeled GNN, L-HGNN (for HGNN in the lorentz space) and P-HGNN (for HGNN in the poincaré space) -- achieve respectable test F1 and accuracy scores. 

Augmenting the models with the POS-tag embeddings uniformly boosts performance across all architectures. Specifically, we observe increments in the test F1 and accuracy scores for all three models. Notably, P-HGNN-POS achieves both our best overall test F1 and accuracy scores of \textbf{84\%} and \textbf{92.1\%} respectively, beating the best test accuracy reported in Table~\ref{tab:baselines} (91.7\%) and coming very close to the best test F1 score (84.9\%), both achieved by their largest model RoBERTa\textsubscript{large} consisting of \textbf{355 million parameters}. Both GNN-POS and L-HGNN-POS also show competitive test F1 scores of \textbf{78.5\%} and \textbf{79.4\%} respectively, while achieving near SOTA accuracy scores of \textbf{89.1\%} and \textbf{89.4\%}. 

Next, we address the imbalance in the dataset through the introduction of a weighted loss function. The Balanced-GNN improves the test F1-score by over 2 points compared to the standard GNN (from \textbf{75.0\%} to \textbf{77.2\%}), demonstrating the effectiveness of the weighted loss for the Euclidean model. The impact on the hyperbolic models is more nuanced, with slight shifts in the precision-recall trade-off resulting in minor changes to the F1-score. In Table~\ref{tab:weight_balance}, we show the best GNN configurations taken from Table~\ref{tab:results} along with all their corresponding weight-balanced versions trained with the same dropout rates. While the early stopping criterion favors the best test F1-score during training, we can still observe the generally expected trend of dropping precision and increasing recall for both the dev and test sets when applying class weights. Interestingly, the Balanced-L-HGNN model does not always follow this pattern as strictly as its euclidean or poincaré counterparts. 

Finally, the models incorporating all enhancements -- POS embeddings and class balancing (-POS-Balanced) -- demonstrate the most overall robust performances, effectively addressing both feature representation and data imbalance. The Balanced-GNN-POS model achieves a strong test F1 and accuracy scores of \textbf{80.0\%} and \textbf{89.4\%}, a clear improvement over its unbalanced version with test F1 and accuracy scores of \textbf{78.5\%} and \textbf{89.1\%}. Most significantly, while the P-HGNN-POS model achieves our highest test F1-score of \textbf{84.0\%}, the Balanced-P-HGNN-POS model achieves a competitive F1-score \textbf{82.6\%} while substantially boosting test recall from \textbf{82.1\%} to \textbf{85.1\%} which is our best test recall score after P-HGNN (\textbf{86.6\%}) and also achieving the best test accuracy of \textbf{90.9\%} after P-HGNN-POS (\textbf{92.1\%}).

Furthermore, it is worth noting the consistently high AUC-ROC scores across all our model configurations, as detailed in Table~\ref{tab:results}. The test set AUC-ROC values range from 0.93 to 0.95, indicating a strong ability of the models to distinguish between the `Claim' and `Not Claim' classes. This high level of class separability further reinforces the reliability of our graph-based approach for the task of environmental claim detection.

The key advantage of our approach lies in its computational efficiency. In Table~\ref{tab:model_params}, we detail the parameter counts for all~\cite{stammbach2022environmental} transformer models as well as our own GNN models. While RoBERTa\textsubscript{large}, the best performing model for environmental claim detection from~\cite{stammbach2022environmental} consists of \textbf{355 million parameters}, our GNN and HGNN models are significantly more lightweight. Our models consist of 4 GNN layers, a 256-dimensional hidden state, and 45 unique dependency relations (edge types). We calculate the size of our graph models to be approximately \textbf{12M parameters}, nearly \textbf{30 times smaller} than RoBERTa\textsubscript{large}~\footnote{\label{common_foot}See Appendix for more details.}. \textit{Therefore, we conclude that our graph-based models achieve better than or comparable to SOTA results.}

\begin{table*}[!h]
\centering
\scriptsize
\begin{tabular}{lcc|ccccc|ccccc}
\toprule
\textbf{Model} & \multicolumn{2}{c|}{\textbf{grid-search parameters}} & \multicolumn{5}{c|}{\textbf{dev}} & \multicolumn{5}{c}{\textbf{test}} \\
 & \makecell{Dropout \\ Rate} & \makecell{Class \\ Weights} & pr & rc & F1 & acc & auc & pr & rc & F1 & acc & auc \\
\midrule
GNN
& 0.1 & --
& \textbf{79.3} & 69.7 & 74.2 & \textbf{87.9} & \textbf{0.93} & \underline{78.7} & 71.6 & \underline{75.0} & \underline{87.9} & \textbf{0.93} \\
Balanced-GNN 
& 0.1 & [0.6678,1.9897]
& {68.3} & \textbf{84.8} & \underline{75.7} & 86.4 & \textbf{0.93} & {69.2} & \textbf{80.6} & 74.5 & 86.0 & \textbf{0.93} \\
Balanced-GNN 
& 0.1 & [0.8,1.6]
& {72.4} & \underline{83.3} & \textbf{77.5} & \textbf{87.9} & \textbf{0.93} & {70.1} & \textbf{80.6} & \underline{75.0} & 86.4 & \textbf{0.93} \\
Balanced-GNN 
& 0.1 & [1,1.5]
& \underline{78.6} & 66.7 & 72.1 & \underline{87.2} & \textbf{0.93} & \textbf{81.7} & \underline{73.1} & \textbf{77.2} & \textbf{89.1} & \textbf{0.93} \\
\midrule
L-HGNN 
& 0.1 & --
& 70.3 & \underline{78.8} & 74.3 & 86.4 & \underline{0.92} & \underline{73.7} & \textbf{83.6} & \textbf{78.3} & \textbf{88.3} & \textbf{0.93} \\
Balanced-L-HGNN 
& 0.1 & [0.6678,1.9897]
& 71.6 & \textbf{80.3} & \underline{75.7} & {87.2} & \textbf{0.93} & 68.8 & \underline{82.1} & 74.8 & 86.0 & \textbf{0.93} \\
Balanced-L-HGNN 
& 0.1 & [0.8,1.6]
& \underline{75.7} & \textbf{80.3} & \textbf{77.9} & \textbf{88.7} & \textbf{0.93} & 73.0 & 80.6 & \underline{76.6} & \underline{87.5} & \textbf{0.93} \\
Balanced-L-HGNN 
& 0.1 & [1,1.5]
& \textbf{79.7} & 71.2 & {75.2} & \underline{88.3} & \textbf{0.93} & \textbf{75.4} & 73.1 & 74.2 & 87.2 & \textbf{0.93} \\
\midrule
P-HGNN 
& 0 & --
& \textbf{71.0} & 74.2 & 72.6 & \underline{86.0} & \underline{0.91} & \textbf{74.4} & \textbf{86.6} & \textbf{80.0} & \textbf{89.1} & \textbf{0.94} \\
Balanced-P-HGNN 
& 0 & [0.6678,1.9897]
& 69.1 & \textbf{84.8} & \textbf{76.2} & \textbf{86.8} & \textbf{0.92} & \underline{71.8} & 83.6 & 77.2 & \underline{87.5} & \underline{0.93} \\
Balanced-P-HGNN 
& 0 & [0.8,1.6]
& \underline{69.9} & {77.3} & \underline{73.4} & \underline{86.0} & \underline{0.91} & 68.2 & \textbf{86.6} & 76.3 & 86.4 & \underline{0.93} \\
Balanced-P-HGNN 
& 0 & [1,1.5]
& 68.4 & \underline{78.8} & 73.2 & 85.7 & \textbf{0.92} & 71.2 & \underline{85.1} & \underline{77.6} & \underline{87.5} & \underline{0.93} \\

\bottomrule
\end{tabular}
\caption{Results for all weight-balanced GNNs. For the best base GNN configurations, we show all the corresponding weight-balanced GNNs at the same dropout rate as the base model. For each model type, i.e., GNN, L-HGNN and P-HGNN, the best performance per split is indicated block-wise in bold, while the second best in underlined.}
\label{tab:weight_balance}
\end{table*}

\begin{table}[!h]
\centering
\normalsize
\begin{tabular}{c|c}
\toprule
\textbf{Model} & \textbf{Parameter-count} \\
\toprule
DistilBERT & 66m \\
ClimateBERT & 82m \\
RoBERTa\textsubscript{base} & 125m \\
RoBERTa\textsubscript{large} & \textbf{355m} \\
GNN/HGNN & \underline{12m} \\
GNN-POS/HGNN-POS & \underline{12m} \\
\bottomrule
\end{tabular}
\caption{Number of parameters for the transformer models used by~\cite{stammbach2022environmental} compared to our GNN and HGNN models. Models prefixes are dropped since they do not affect the parameter sizes. m stands for million. Largest model size is in bold while the smallest is underlined.}
\label{tab:model_params}
\end{table}

\subsection{HGNNs Consistently Outperform GNNs (RQ2.)}
In Table~\ref{tab:results}, we observe that hyperbolic GNN models, particularly those in the poincaré space consistently outperform their euclidean counterparts under most configurations for both the F1 and accuracy scores. For example on the test set, both the L-HGNN with an F1-score of \textbf{78.3\%} and accuracy of \textbf{88.3\%} as well as the P-HGNN with an F1-score of \textbf{80.0\%} and an accuracy of \textbf{89.1\%} surpass the standard GNN with an F1 score of \textbf{75.0\%} and accuracy of \textbf{87.9\%}. Similarly, this trend is continued in other configurations and the performance gap widens with the inclusion of richer features, as seen with P-HGNN-POS (\textbf{84\%} F1 and \textbf{92.1\%} accuracy) outperforming GNN-POS (\textbf{78.5\%} F1 and \textbf{89.1\%} accuracy). This consistent advantage shows that hyperbolic space models significantly benefit from explicit hierarchical modeling of the data using tree-like structures such as dependency parsing graphs. We achieve better test scores with HGNNs
than with GNNs under most configurations, indicat-
ing a low hyperbolicity (i.e., a strong hierarchical
structure) in the ECD dataset. \textit{Therefore, we conclude that explicit hierarchical modeling of environmental claims allows the geometric properties of hyperbolic models to benefit from this hierarchy and improve over their euclidean counterparts.}

%% file: latex/discussion.tex
\section{Discussion}

In this study, we investigate the efficacy of Graph Neural Networks (GNNs) and their hyperbolic counterparts (HGNNs) for Environmental Claim Detection. We construct dependency parsing graphs of claim sentences to explicitly model them as hierarchical structures, hence benefiting from the geometric properties of the hyperbolic space. Leveraging simple word embeddings for node features, we also incorporate POS-tags and a weighted loss function to enhance performance and address data imbalance.

Our results indicate that graph-based models, particularly those in the hyperbolic space, can achieve performance superior to SOTA transformer-based architectures. The P-HGNN-POS model, our best-performing configuration, achieves a test F1-score of \textbf{84.0\%} and an accuracy of \textbf{92.1\%}, even surpassing the \textbf{91.7\%} accuracy of the much larger RoBERTa\textsubscript{large} model. This performance is achieved with approximately \textbf{12 million parameters}, a nearly 30-fold reduction compared to the \textbf{355 million parameters} of RoBERTa\textsubscript{large}. These findings highlight the potential of graph-based models as lightweight, efficient, and effective alternatives to LLMs for specialized NLP tasks.

\noindent\textit{\textbf{Takeaway for RQ.1}}: Graph-based models offer a computationally efficient alternative to large transformers for environmental claim detection without compromising performance.

Furthermore, our results demonstrate the potential of hyperbolic geometry for NLP tasks like claim detection. Across various configurations, HGNNs consistently outperform GNNs, and this performance gap becomes more pronounced with the introduction of richer syntactic features, as seen in the superior performance of P-HGNN-POS over GNN-POS. This suggests that tree-like modeling of sentence structure creates a hierarchical representation that is naturally well-suited to the geometric properties of hyperbolic space.

\noindent\textbf{\textit{Takeaway for RQ.2}}: Explicit hierarchical modeling of claims significantly benefits hyperbolic models, indicating their potential for NLP tasks.

Our findings underscore two critical points for the field. First, the dominance of transformer-based models is not absolute; for specific, well-defined tasks like environmental claim detection, specialized and lightweight models like GNNs can provide more efficient and effective solutions. Second, the inherent, often implicit, hierarchical nature of linguistic data can be powerfully exploited by choosing geometric spaces -- like hyperbolic space -- that align with this underlying structure. This highlights the vast potential in exploring geometries beyond euclidean for learning efficient representations for NLP tasks.

%% file: latex/conclusion.tex
\section{Conclusion}
In this work, we introduce an efficient graph-based methodology for environmental claim detection, positioning GNNs and HGNNs as lightweight yet effective alternatives to transformer-based architectures. We reformulate the task as a graph-classification problem, transforming claim sentences into dependency parsing graphs with simple word and POS-tag embeddings as node features and encoding syntactic dependencies as edge relations. Our results demonstrate that GNNs achieve performance comparable or superior to SOTA models with a 30-fold reduction in parameters. Furthermore, we consistently observe that HGNNs outperform their GNNs, affirming that the geometric properties of HGNNs gain significant advantage from the explicit hierarchical modeling of the data. Our findings call for a shift beyond over-reliance on transformers, demonstrating that specialized models can yield more efficient solutions for targeted NLP tasks without a loss of capability.

\noindent\textbf{Future work.} First, we plan to compare static word2vec embeddings for node features with sentence embeddings from transformer models like RoBERTa. Second, we plan to move beyond simple one-hot encoded edge features to a knowledge-enhanced schema based on principles from universal dependencies~\cite{de2021universal}. Third, we plan to experiment with alternative graph representations beyond dependency parsing such as constituency parsing. Fourth, we plan to conduct a sensitivity analysis to quantify the impact of parsing inaccuracies on model performance, investigating whether domain-adapted parsers could yield better results. Lastly, to assess the generalisability of this study, we intend to extend our work to more NLP tasks, models such as graph attention networks~\cite{velivckovic2017graph}, and benchmark datasets such as FEVER~\cite{thorne2018fever} and Climate-Fever~\cite{diggelmann2020climate}.

%% file: latex/limitations.tex
% \clearpage
\section{Limitations}

% Our study highlights the potential of GNNs and HGNNs for environmental claim detection, but there are several limitations to our methodology and the broader application of graph structures in sequential modeling tasks.

% \begin{itemize}

%     \item \textbf{Limitations of Graph Structures in Sequential Modeling}: While graph-based approaches offer a novel way to model hierarchical and relational structures, most textual data is inherently sequential and better suited to traditional models like transformers. Adapting sequential data for tasks like graph classification is complex and often unintuitive, requiring significant preprocessing and domain-specific adaptations that may not generalize. As a result, while graphs are a powerful tool to model certain NLP applications, they may struggle with the broader range of sequential tasks where traditional models excel.
% \end{itemize}

We highlight the limitations of our work as follows.
\begin{compactitem}
    \item Our approach deliberately utilizes word2vec embeddings for node features to create a maximally lightweight and efficient model. However, they do not encode the sequential dependencies between words. Similarly, our edge features are simple one-hot encodings of dependency types, which treat all syntactic relations as independent and do not capture potential similarities between them.
    \item Our methodology relies on the output of the dependency parser to construct the graphs. While modern parsers are highly accurate, any errors in graph construction are propagated as noise to the GNN and HGNN models. We do not analyze the impact of such parsing errors on final model performance in this study.
    \item The scope of our experiments is focused on a single, relatively small, English-only dataset. While the results are strong, the generalisability of our graph-based approach to other claim detection domains, larger datasets, and other languages is yet to be established.
    \item The transformer baselines used for comparison are from the original environmental claim detection paper~\cite{stammbach2022environmental}. We do not benchmark our models against more recent, state-of-the-art LLMs such as Llama3~\cite{dubey2024llama} and GPT-4o~\cite{hurst2024gpt}, limiting the assessment of our approach against the current SOTA.
\end{compactitem}

%% file: latex/supplementary.tex
\section{Appendix}

\begin{figure*}[ht]
    \centering
    % Dependency Graph
    \begin{dependency}[theme = simple]
       \begin{deptext}[column sep=0.3cm]
          Gas \& is \& also \& a \& cleaner \& fuel \& with \& resultant \& environmental \& benefits \\
          \small\textit{NOUN} \& \small\textit{AUX} \& \small\textit{ADV} \& \small\textit{DET} \& \small\textit{ADJ} \& \small\textit{NOUN} \& \small\textit{ADP} \& \small\textit{ADJ} \& \small\textit{ADJ} \& \small\textit{NOUN} \\
       \end{deptext}
       \deproot{2}{ROOT}
       \depedge{2}{1}{nsubj}
       \depedge{2}{3}{advmod}
       \depedge{2}{6}{attr}
       \depedge[edge below]{6}{4}{det}
       \depedge{6}{5}{amod}
       \depedge{6}{7}{prep}
       \depedge{7}{10}{pobj}
       \depedge[edge below]{10}{8}{amod}
       \depedge{10}{9}{amod}
    \end{dependency}
    \caption{The transformation of the example claim into a dependency graph. The graph shows tokens and their POS tags as nodes, with syntactic dependencies as labeled, directed edges.}
    \label{fig:dep_graph_example}
\end{figure*}

\subsection{Generating Dependency Parsing Graphs of Environmental Claims}
We now provide a working example of the process of converting claim sentences into their dependency parsing graphs. The feature vectors shown are for illustrative purposes and do not represent actual embedding values. Let the claim sentence $\mathbf{C}$ be \textit{``Gas is also a cleaner fuel with resultant environmental benefits.''} 
% Node and Edge Feature Explanation
    \begin{compactitem}
        \item\textbf{Dependency Parsing $\mathbf{C}$: }The claim sentence $\mathbf{C}$ is first transformed into its corresponding dependency parsing graph using the spaCy dependency parser. Figure~\ref{fig:dep_graph_example} illustrates this transformation.
        \item\textbf{Node Features ($x'_{v}$):} Each node's feature vector is the concatenation of its word embedding and a randomly initialized, trainable POS tag embedding. For the node `cleaner` (an `ADJ`), with $d_{word}=4$ and $d_{pos}=2$,
        \[
        x'_{\text{cleaner}} = [\underbrace{W_{e}(\text{`cleaner'})}_{\text{word2vec}} \mathbin\Vert \underbrace{W_{p}(\text{`ADJ'})}_{\text{POS emb.}}] 
        \]
        \[
        = \left[ \begin{pmatrix} 0.21 \\ -0.45 \\ 0.67 \\ 0.09 \end{pmatrix} \middle\Vert \begin{pmatrix} 0.62 \\ 0.15 \end{pmatrix} \right] = \begin{pmatrix} 0.21 \\ -0.45 \\ 0.67 \\ 0.09 \\ 0.62 \\ 0.15 \end{pmatrix}
        \]
        
        \item\textbf{Edge Features ($e_{hj}$):} Each dependency relation is one-hot encoded. The \textit{amod} relation from `fuel` to `cleaner`, being the 5th unique relation out of 45, is represented as:
        $ e_{\text{fuel, cleaner}} = \begin{pmatrix} 0 & 0 & 0 & 0 & 1 & \cdots & 0 \end{pmatrix}^T \in \mathbb{R}^{45} $
    \end{compactitem}

\subsection{Training Configuration}
\begin{table}[!h]
    \centering
    \setlength{\tabcolsep}{3pt}
    \small
    \begin{tabular}{p{4cm}p{3cm}}
    \toprule
        \textbf{Hyperparameter} & \textbf{Value} \\
            \midrule
            Number of GNN Layers & 4 \\
            Learning Rate & 0.001 \\
            Hyperbolic Learning Rate & 0.001 \\
            Patience (Early Stopping) & 8 \\
            Activation Function & Leaky ReLU \\
            Leaky ReLU Slope & 0.5 \\
            Optimizer & AMSGrad \\
            Hyperbolic Optimizer & Riemannian AMSGrad \\
            Embedding Dimension & 256 \\
            Number of Centroids & 30 \\
            Maximum Epochs & 30 \\
            Edge Types & 45 \\
            Number of Classes & 2 \\
            Initialization Method & Xavier \\
            Gradient Clipping & 1.0 \\
            \bottomrule
            \end{tabular}
            \caption{GNN and HGNN Training Configuration}
  \label{tab:trainConfig} 
\end{table}

\noindent Table~\ref{tab:trainConfig} shows the (fixed) hyperparameters used for training our GNN and HGNN models. Dropout rate, POS-embedding dimension and class-weights were optimized through grid-search.

\subsection{Evaluation Metrics}
Let TP, FP, TN, and FN be the number of True Positives, False Positives, True Negatives, and False Negatives, respectively. The metrics are defined as follows.

\begin{compactitem}
    \item \textbf{Accuracy:} The proportion of correctly classified instances among the total instances.
    $$
    \text{Accuracy} = \frac{TP + TN}{TP + TN + FP + FN}
    $$

    \item \textbf{Precision:} The ratio of correctly predicted positive observations to the total predicted positive observations.
    $$
    \text{Precision} = \frac{TP}{TP + FP}
    $$

    \item \textbf{Recall:} The ratio of correctly predicted positive observations to all observations in the actual class.
    $$
    \text{Recall} = \frac{TP}{TP + FN}
    $$

    \item \textbf{F1-Score:} The harmonic mean of Precision and Recall.
    $$
    \text{F1-Score} = 2 \cdot \frac{\text{Precision} \cdot \text{Recall}}{\text{Precision} + \text{Recall}}
    $$

    \item \textbf{AUC-ROC:} The Area Under the Receiver Operating Characteristic Curve. It measures the model's ability to distinguish between positive and negative classes across all classification thresholds.
\end{compactitem}

\subsection{Parameter Size Calculation for Graph Models}
Here, we detail the number of trainable parameters for our graph models. We first calculate the number of trainable parameters for the base model (with only word embeddings) and then for the model augmented with POS-tag embeddings (-POS).

\noindent\textbf{Base GNN/HGNN Model (without POS).} The base model's parameters are distributed across an input projection layer, three hidden GNN layers, and a final classifier.

\begin{itemize}
    \item Layer 1 (Input Projection): Maps the 300-dimensional word embeddings to the 256-dimensional hidden space for each of the 45 relations. 45 relations $\times$ 300 input\_dim $\times$ 256 output\_dim) + 256 bias = 3,456,256.
    \item Layers 2, 3, \& 4: These three layers map the 256-dimensional hidden state to another 256-dimensional hidden state for each relation. 3 layers $\times$ [(45 relations $\times$ 256 input\_dim $\times$ 256 output\_dim) + 256 bias] = 8,848,128.
    \item Final Classifier: (256 $\times$ 2 output classes) + 2 bias = 514.
    \item Total number of parameters = 3,456,256 + 8,848,128 + 514 = \textbf{12,304,898}.
\end{itemize}

\noindent \textbf{GNN-POS/HGNN-POS Model (with POS).} Adds learnable POS embeddings. Using an example POS dimension (d\textsubscript{pos}) be 16:

\begin{itemize}
    \item POS Tag Embeddings: 18 vocab\_size $\times$ 16 pos\_dim = 288
    \item Layer 1 (Input Projection): The input dimension is now 300 + 16 = 316. (45 $\times$ 316 $\times$ 256) + 256 = 3,640,576.
    \item Layers 2, 3, \& 4: Unchanged from the base model. Parameters: 8,848,128
    \item Final Classifier: Unchanged from the base model. Parameters: 514
    \item Total number of parameters (POS) = 288 + 3,640,576 + 8,848,128 + 514 = \textbf{12,489,506}.
\end{itemize}